\pdfoutput=1

\documentclass[11pt]{article}

\usepackage[]{ACL2023}

\usepackage{times}
\usepackage{latexsym}
\usepackage{amsthm,amsmath,amssymb} 
\usepackage{mathrsfs} 
\usepackage{bm} 
\usepackage{pifont} 
\usepackage[T1]{fontenc}
\usepackage{float}

\usepackage[utf8]{inputenc}

\usepackage{microtype}

\usepackage{inconsolata}

\usepackage{diagbox}
\usepackage{multirow}
\usepackage{booktabs}
\usepackage{float}
\usepackage{hyperref}
\usepackage{paralist}
\usepackage{enumitem}

\usepackage[normalem]{ulem}
\useunder{\uline}{\ul}{}

\usepackage[spacesep,definevectors]{easyvector} 
\usepackage{graphicx}
\newcommand{\heading}[1]{\noindent\textbf{#1}\hspace{0.5em}}

\title{Exploiting Correlations Between Contexts and Definitions with Multiple Definition Modeling}

\author{Linhan Zhang$^{1}$ \quad Qian Chen$^{2}$ \quad Wen Wang$^{2}$ \\ 
\textbf{\quad Yuxin Jiang$^{4}$  \quad Bing Li$^{3}$ \quad Wei Wang$^{4}$ \quad Xin Cao$^{1}$}\\
$^1$School of Computer Science and Engineering, The University of New South Wales \\
$^2$Speech Lab, Alibaba Group, China\\
$^3$Institute of High Performance Computing (IHPC), Agency for Science, \\ Technology \& Research (A*STAR), Singapore \\
$^4$Hong Kong University of Science and Technology (Guangzhou), China \\
{\texttt \{linahan.zhang, xin.cao\}@unsw.edu.au } \\
{\texttt \{tanqing.cq,w.wang,\}@alibaba-inc.com}\\
}

\begin{document}
\maketitle
\begin{abstract}

Definition modeling is an important task in advanced natural languages applications such as understanding and conversation. 
Since its introduction, it focus on generating one definition for a target word or phrase in a given context, which we refer to as Single Definition Modeling (SDM). However, this approach does not adequately model the correlations and patterns among different contexts and definitions of words. In addition, the creation of a training dataset for SDM requires significant human expertise and effort. 
In this paper, we carefully design a new task called Multiple Definition Modeling (MDM) that pool together all contexts and definition of target words. We demonstrate the ease of creating a model as well as multiple training sets automatically. %
In the experiments, we demonstrate and analyze the benefits of MDM, including improving SDM's performance by using MDM as the pretraining task and its comparable performance in the zero-shot setting.

\end{abstract}

\section{Introduction}
\label{sec:introduction}
The definition modeling task, first proposed by~\cite{DBLP:conf/aaai/NorasetLBD17}, aims to generate definitions for words. Definition modeling has wide applications, including natural language understanding (NLU) \cite{DBLP:conf/acl/nlu1, DBLP:journals/corr/nlu2} and human-machine conversation \citep{DBLP:journals/ijsc/conv1,DBLP:journals/corr/conv2}.
Despite many works on solving the definition modeling problems~\citep{DBLP:conf/acl/GadetskyYV18,DBLP:conf/naacl/IshiwatariH0NST19, DBLP:conf/emnlp/BevilacquaMN20, DBLP:conf/acl/KongCZYY22}, \emph{all} existing approaches use the following problem formulation where the input consists of a \emph{single} context and the model outputs its corresponding definition. We refer to this formalism as ``Single Definition Modeling'' (SDM). %

\begin{table}[t]
\centering
\scalebox{0.95}{
  \begin{tabular}{|p{10mm}||p{7mm}|p{50mm}|}
    \toprule
       Word  & PoS & Sense/Definition \\\midrule
    spout   & n & a newly \underline{grown} \underline{bud} (especially from a \underline{germinating} seed)      \\\cline{2-3}
            & v & \underline{produce} \underline{buds}, branches, or \underline{germinate} \\\bottomrule
  \end{tabular}
}
\caption{An example of correlations among multiple contexts and definitions of the same word. Underlined words denote the same or semantically related words across definitions.}
\label{tab:correlation}
\end{table}

SDM has several limitations: (1) It misses the opportunity to exploit the correlations among multiple contexts and definitions for the same word. We show a typical example in Table~\ref{tab:correlation} where it is obviously beneficial to learn the highly correlated and semantically related multiple definitions of the same word together. %
(2) The SDM task requires a high-quality training dataset, leading to significant human effort to create one. To select a definition for a given context, users essentially perform word sense disambiguation. Furthermore, for a novel sense, it requires users to create an appropriate context; both require linguistic expertise and patience. 

The above limitations are exacerbated by the fact that languages are constantly evolving and semantic changes have sped up, especially in user-generated text~\cite{jatowta2021computational}. As new words, novel senses, and sense shifts occur, it is time-consuming, laborious, and error-prone to update definitions and usage examples by lexicographers, leading to laggy updates and incompleteness for sense inventories \cite{DBLP:conf/acl/KongCZYY22, DBLP:conf/emnlp/HuangKA21}.



In this paper, we introduce a novel formalism for definition modeling called ``Multiple Definition Modeling'' (MDM) to address the above limitations. In a nutshell, MDM takes as input many contexts of a word and outputs possibly more than one definition. In addition, it does not require the input contexts and output definitions to align with each other.  
As SDM is a special case of MDM, existing SDM models can be easily adapted to solve the MDM task. Furthermore, MDM does not require alignment of the input contexts and output definitions, offering the opportunity to create its training datasets automatically and exploiting open-domain Web texts to generate contexts.  
We showcase the above two features concretely in this paper and conducted extensive experiments to demonstrate the performance of MDM tasks and datasets in several different settings. 
%


The main contributions of our work are:
\begin{itemize}[leftmargin=*,noitemsep]
\item We introduce a new task called Multiple Definition Modeling (MDM) for generating multiple definitions of a word based on all of its contexts in a dataset.  
\item Thanks to the careful design of the MDM task, we demonstrate the ease of creating a model for MDM and creating a new dataset (WordWiki) for MDM training, fully automatically using open-domain Web text.
\item We conduct experiments to showcase several benefits of the MDM model in comparison to its SDM counterpart, including outperforming SDM in a zero-shot setting, comparable performance with SDM in an out-of-domain setting, and promising results on polysemous words. 
\end{itemize}

\section{The MDM Task Formulation}
\label{sec:mdm_formulation}

In view of the limitations of SDM elaborated in Section~\ref{sec:introduction}, we propose the novel task formulation termed \textbf{Multiple Definitions Modeling (MDM)}. Taking multiple contexts of a word as input, MDM requires a model to learn how to simultaneously predict multiple definitions of the word given these contexts. Concretely, given a target word $w$, its MDM input consists of \emph{all} contexts (i.e., sentences) $C=\{c_i\}_{i=1}^{N}$ where $w$ appears each $c_i$ in a dataset; its MDM output is \emph{all} gold definitions $G=\{g_i\}_{i=1}^{M}$ for $w$ in a sense inventory (e.g., WordNet). 

MDM is specially designed so that it facilitates creating its training datasets \emph{automatically} from \emph{open-domain Web} texts, as it does not require:
\begin{inparaenum}[(1)]
    \item input contexts and output definitions to align with each other in order, hence in general $N \neq M$; and
    \item the context to be carefully selected, hence allowing the contexts $C$ of $w$ to be created automatically from abundant sentences containing $w$ on the Web. 
\end{inparaenum}

As SDM is a special case of MDM, we could use a model trained on the MDM as a good initialization for SDM training and hence benefit any SDM models and applications. Moreover, benefiting from learning correlations among definitions for multiple senses and the benefit of generating \emph{multiple} definitions,  MDM could capture the implicit common generative patterns among contexts and word definitions and perform well when there is a novel sense outside the training dataset, e.g., as it is typical in the zero-shot settings.

\section{Models for MDM Tasks}
\label{subsec:definition_sum}

In this section, we illustrate the ease of creating a model for the MDM task. A popular model for the SDM task is based on the text-to-text model, such as T5~\cite{raffel2020exploring} or BART~\cite{DBLP:conf/acl/LewisLGGMLSZ20}. %
It can be easily adapted to tackle the MDM task. Specifically,  we concatenate the input contexts $C=\{c_i\}_{i=1}^{N}$ with the  
delimiter \verb!<sep>! to form the input token sequence denoted as $x$. 
Similarly, the gold definitions for $w$, i.e., $D=\{d_i\}_{i}^M$ are also concatenated by \verb!<sep>!,  denoted as $y$. This simple reduction procedure effectively turns the MDM inputs/outputs into SDM-compatible ones. 

The objective of the MDM model is then the same as that of the SDM model, which is to maximize the probability of the generated definition sequence $y$ conditioned on its context sequence $x$. All model enhancement techniques can be used. E.g., to reduce the difficulty of training with a transformer-based model, the actual token sequence fed into the model is prefixed with their roles, i.e., as ``\texttt{word: $w$ context: $x$}''. For a pair $(x,y)$ in the training set $S$, the probability of $y$ is calculated autoregressively as:
\begin{equation}
\small
    P(\tilde{y} \mid x, \Btheta) = \quad\prod_{t=1} P( \tilde{y_{t}} \mid y_{<t},x;\Btheta) \quad 
\end{equation}
\noindent where $\tilde{y}$ is the predicted definition sequence and  $\tilde{y_{t}}$ is the $t$-th token in $\tilde{y}$. The weights $\Btheta$ of the model are optimized by the cross-entropy loss:
\begin{equation}
\small
    \mathcal{L}_{sum}(y, \tilde{y}) = -\sum_{(x,y) \in S} \log P( \tilde{y} \mid x;\Btheta) 
\end{equation}

\section{Training Datasets for MDM Tasks}
\label{sec:wordwiki}


In this section, we demonstrate the ease of creating a training dataset automatically for the MDM task. Concretely, we describe the details of creating the \textbf{WordWiki} dataset. 

We utilize an existing SDM dataset Wordnet for its glosses and the definitions and utilize the open-domain Wikitext\footnote{https://huggingface.co/datasets/wikitext} to create the contexts. Concretely, we first employ the \emph{NLTK} toolkit\footnote{https://www.nltk.org/} to tokenize documents in Wikitext into words in lowercase. Next, we remove infrequent words as they are mostly garbled characters or meaningless words, as well as words not in the Wordnet glosses. 
For each remaining word $w$, we look up its definition(s) in Wordnet and format them as golden labels. We then obtain $N$ sentences in Wikitext that contain $w$. %
We can create multiple training datasets with different difficulty levels by choosing $N = M + k$, where $k \in \{0,2,4\}$. $M$ is the number of definitions. Hence, a larger $k$ typically provides more contexts for the model to learn accurately and sufficiently. %
To cope with the maximum sequence length limitation of the T5-base model, we randomly sample $N$ contexts if $w$ occurs in many more sentences and use truncation if necessary.  
%
%
The statistics of WordWiki are shown in Table~\ref{table_wordwiki}.




\begin{table}[htpb]
\centering
\scalebox{0.8}{
\begin{tabular}{@{}c|c|c|c@{}}
\toprule[1.5pt]
DATASET                      & Entry number          & K   & \#Word number \\ \midrule
MDM-WordNet                       & 7938                  & N/A & 5.81        \\ \midrule
\multirow{3}{*}{WordWiki} & \multirow{3}{*}{7543} & 0   & 58.73       \\ \cmidrule(l){3-4} 
                          &                       & 2   & 127.06      \\ \cmidrule(l){3-4} 
                          &                       & 4   & 194.33      \\  \bottomrule[1.5pt]
\end{tabular}
}
\caption{Data statistics of \textbf{MDM-WordNet} and \textbf{WordWiki}, including the total number of entries and also the number of \#words in contexts for train sets.}
\label{table_wordwiki}
\end{table}
\vspace{-6mm}
\section{Experiments}
\label{sec:experiment}
\subsection{Datasets}
\label{subsec: sdm-dataset}

We used the following datasets in the experiments. 

\heading{SDM-WordNet} SDM benchmarks are used for evaluating the performance of using MDM models as pre-trained models on the SDM task. For fair comparisons, we used WordNet~\cite{DBLP:conf/naacl/IshiwatariH0NST19} as SDM benchmark for evaluation (denoted as \textbf{SDM-WordNet} in this paper). Each entry in WordNet comprises a word and one of its senses with the corresponding usage example. The usage example and its definition were used as a source and target for definition generation models. The details of WordNet could refer to ~\cite{DBLP:conf/naacl/IshiwatariH0NST19}. We followed previous work \cite{DBLP:conf/emnlp/HuangKA21} to clean noisy entries from the dataset.   

\heading{MDM-WordNet}  We also utilize WordNet to construct a MDM dataset for the \emph{MDM-Easy} task (See Section~\ref{sec:wordwiki}), with the difference that each definition is aligned with its context, which means the order of definitions should be in line with that of contexts; the resulting dataset is denoted as \textbf{MDM-WordNet}. The settings for \textbf{MDM-Easy} task is the contexts and definitions align with each other (hence $N = M$) while \textbf{MDM-Hard} has no restriction. 

\heading{MDM-WordWiki} We adopt the dataset introduced in Section \ref{sec:wordwiki} to conduct experiments for \emph{MDM-Hard} task. Different from \emph{MDM-Easy} task, \emph{MDM-Hard} has no restrictions on the alignment of contexts and definitions.

\subsection{Experimental Setup and Results}
\label{subsec:experimental-setup}

\begin{table}[]
\centering
\scalebox{0.8}{
\begin{tabular}{@{}l|cc|c@{}}
\toprule[1.5pt]
\textbf{Experiment} & \multicolumn{2}{c|}{\textbf{Model}}                          & \textbf{WordNet(BLEU)} \\ \midrule
\multirow{3}{*}{1}  & \multicolumn{2}{c|}{T5-base(ours)}                           & 30.34                  \\ \cmidrule(l){2-4} 
 & \multicolumn{2}{c|}{T5-base MDM-Easy P1} & 30.77 \\ \cmidrule(l){2-4} 
 & \multicolumn{2}{c|}{T5-base MDM-Easy P2} & 31.27 \\ \midrule[1.5pt]
\multirow{3}{*}{2}  & \multicolumn{1}{c|}{\multirow{3}{*}{T5-base MDM-Hard}} & k=0 & 28.53                  \\ \cmidrule(l){3-4} 
 & \multicolumn{1}{c|}{}        & k=2       & 28.56 \\ \cmidrule(l){3-4} 
 & \multicolumn{1}{c|}{}        & k=4       & 28.78 \\ \bottomrule[1.5pt]
\end{tabular}
}%
\caption{MDM-Easy P1 and MDM-Easy P2 means Phase 1 and Phase 2. Phase1 is trained by MDM-WordNet. Phase 2 is models trained on Phase 1 and then finetuned on SDM-WordNet dataset. MDM-Hard utilizes WordWiki datasets. Besides, as described in \ref{sec:wordwiki}, k means the number of contexts more than the number of definitions. }
\label{result:exp1}
\end{table}

\begin{table}[]
\centering
\scalebox{0.8}{
\begin{tabular}{@{}l|c|c@{}}
\toprule[1.5pt]
\multicolumn{1}{c|}{\textbf{Model}} & \textbf{Subset Def=1} & \textbf{Subset Def\textgreater{}1} \\ \midrule
T5-base(ours)        & 26.87 & 34.91 \\ \midrule
T5-base MDM-Easy P1  & 27.38 & 35.23 \\ \midrule
T5-base MDM-Easy P2  & 27.24 & 36.25 \\ \midrule[1.2pt]
T5-base MDM-Hard k=0 & 24.88 & 33.34 \\ \midrule
T5-base MDM-Hard k=2 & 24.67 & 33.68 \\ \midrule
T5-base MDM-Hard k=4 & 24.78 & 34.84 \\ \bottomrule[1.5pt]
\end{tabular}
}
\caption{The table presents the results of \textbf{Experiment 3}, where the WordWiki-trained models are tested on different SDM datasets including data with a single definition and with multiple definitions.  Subset Def=1 and Subset Def>1 are test data from SDM-WordNet which contains words with only one definition and multiple definitions.
The evaluation meric is BLEU score and $k$ is as same as the one described in \ref{sec:wordwiki}.}
\label{result:exp2}
\end{table}

In our experiments, we used the Adam optimizer and set the batch size to 16. The total number of training epochs was 140, with 70 epochs per phase for the 2-phase training setup, ensuring comparability across experiments. For all SDM models (including the baseline and MDE-Easy phase 2 in Tables \ref{result:exp1} and \ref{result:exp2}), the learning rate was set to 3e-4. For MDM-Easy phase 1, the learning rate was 1e-4, while it was 2e-4 for MDM-Hard.
We used BLEU~\cite{doddington2002automatic} as the evaluation metric. 

\heading{Experiment 1}%
We demonstrate the benefit of using a MDM as a pretraining task to improve the model's performance on a SDM dataset. We compare the following two methods: 
\begin{inparaenum}[(1)]
\item An SDM model based on the T5-base model\footnote{https://huggingface.co/facebook/T5-base} trained \emph{only} on the SDM-WordNet dataset. 
\item We pre-train a T5-base model in the \emph{MDM-Easy} setting on the MDM-WordNet dataset, and then finetune and evaluate it on the test set of SDM-WordNet. 
\end{inparaenum}
In Table~\ref{result:exp1}, T5-base trained on MDM-WordNet and then finetuned on SDM-WordNet ($31.27$) achieved the better performance compared with T5-base trained on WordNet directly ($30.34$), with an improvement of $0.93$. We also notice that even without fine-tuning, the model pretrained on the MDM dataset ($30.77$)
outperforms the baseline (by $0.43$).

 
\heading{Experiment 2}%
We demonstrate the impact of increasing the number of contexts for each word in the MDM training dataset to the SDM test set performance. Consequently, we switch to the MDM-Hard setting (i.e., the input contexts and output definitions do not need to align with each other). In addition, we use the newly created MDM-WordWiki as the MDM dataset for training --- note that we do not use any SDM dataset for finetuning. 

From Table~\ref{result:exp1}, models trained only on the MDM dataset in the hard setting have a comparable performance with the same model training on the SDM dataset (See Line 2-3 in Table~\ref{result:exp1}). This is significant as (1) the result is an out-of-domain test for our MDM-only model as the training data and testing data are not from the same distribution, and (2) the MDM dataset does not contain an explicit alignment signal for the model to learn. 

We also note that for our model training only on the MDM dataset improves as the value of $k$ increases. This trend demonstrates the benefit of using more and diverse contexts from open-domain sources to improve the performance of definition modeling models.

\heading{Experiment 3} %
We investigate the detailed performance for models trained on the MDM dataset for words with only one definition versus words with multiple definitions. To do this, we divided the test data into two subsets based on the number of definitions. We then used the models trained only on MDM-WordWiki to make predictions on these two subsets in a zero-shot manner.

As we can see from Table~\ref{result:exp2}, there are no noticeable difference among the models trained on MDM-WordWiki with varying $k$ for words with a single definition.  However, as $k$ increases, the performance of the models for polysemous words improves. Specifically, when $k=4$, the BLEU score increases by an absolute $1.5$ compared to $k=0$. This suggests that data augmentation with additional open-domain contexts can indeed enhance the model's performance on polysemous words. Furthermore, the lack of alignment between contexts and definitions does not impede the model's ability to learn from the correlations between different contexts and definitions.  

\begin{table}[t]
\centering
\scalebox{0.95}{
\begin{tabular}{@{}l|ll@{}}
\toprule[1.5pt]
WORD & \multicolumn{2}{l}{\textbf{freakish}}              \\ \midrule
CONTEXT & \multicolumn{2}{l}{a freakish extra toe}   \\ \midrule
DEF  & \multicolumn{2}{l}{characteristic of a freak} \\ \midrule
SDM\_PRED & \multicolumn{2}{l}{\begin{tabular}[c]{@{}l@{}}unusually bad or displeasing \end{tabular}}       \\ \midrule
MDM-PRED  & \multicolumn{2}{l}{\begin{tabular}[c]{@{}l@{}}extremely freaky\end{tabular}} \\ 
  \bottomrule[1.5pt]
  \end{tabular}
}
\caption{An example of predictions from T5-base trained with SDM and MDM task on WordNet dataset.}
\label{tab:case_study}
\end{table}

\heading{Case study}%
Table \ref{tab:case_study} lists one of the predictions from T5-base(ours) and T5-base MDM-Easy P1. Other case studies could be found in Appendix~\ref{sec:appendix}.

\vspace{-3mm}
\section{Conclusion}
\vspace{-2mm}
We propose a new Multiple Definition Modeling (MDM) task to generate multiple definitions of a word simultaneously given all its contexts. We create a new dataset WordWiki to support MDM.
Experiments verify that MDM could effectively address the limitations of single definition modeling (SDM) task. MDM improves context comprehension for models and enables the generation of more accurate and abstractive definitions. Models without fine-grained data could also achieve comparable performance on SDM tasks. 

\newpage
\section{Limitations}
There are some potential limitations of our proposed MDM task. Due to time limitations, we have not explored the augmentation of MDM-WordNet by producing entries where the number of definitions and contexts are unequal. And also, the diversity of predictions of the MDM task is also a problem that the predicted multiple definitions should be semantically related but also be distinguishable. Different from previous methods for the diversity of generation models, the diversity of MDM task enquires keeping the sentence-level diversity instead of just ensuring predicted tokens/n-grams be different from the previously predicted units. Hence, we need to tackle these issues in future works.

\section{Ethical Considerations}
The MDM task is exceptionally challenging. When the generated new out-of-inventory sense is errorful, it could prevent people from correctly understanding this word. So the generated new definitions could be used for assisting understanding, but they cannot fully replace careful manual verifications.


\bibliography{reference}
\bibliographystyle{acl_natbib}

\appendix

\newpage
\section{Example Appendix}
\label{sec:appendix}
\subsection{Related Work}
\label{related_work}
The first study on definition modeling is proposed by \citet{DBLP:conf/aaai/NorasetLBD17}, which uses word embeddings to generate corresponding word definitions by an RNN-based sequence-to-sequence model. 
However, the problem of polysemes cannot be addressed by only exploiting word embedding. 
\citet{DBLP:conf/ijcnlp/NiW17} first attempted to model local contexts of a word for Internet slang explanation generation. Following this work, many works argued utilizing contexts could help generate multiple definitions for polysemes.  \citet{DBLP:conf/acl/GadetskyYV18} employed Adaptive Skip-Gram model \cite{pmlr-v51-bartunov16} to learn the required number of vector representations for each polyseme. 
\citet{DBLP:conf/naacl/IshiwatariH0NST19} created a Wikipedia dataset that takes the descriptions of a word as target and uses the sentences referring to the target item from Wikipedia articles as contexts. 
Nevertheless, \citet{DBLP:conf/acl/GadetskyYV18,chang2018xsense, DBLP:conf/naacl/IshiwatariH0NST19,li-etal-2020-explicit,washio-etal-2019-bridging, chang-chen-2019-word} mainly used definition modeling to obtain static sense embedding from contexts, which did not fully utilize the context information and the sense embeddings are not contextual.

Recently, more and more definition modeling methods explore Transformer \cite{DBLP:journals/corr/VaswaniSPUJGKP17} based pretrained natural language generation models \cite{DBLP:journals/corr/abs-1911-05715, DBLP:conf/acl/LewisLGGMLSZ20,raffel2020exploring, DBLP:journals/corr/abs-1810-04805} to generate definitions given the contexts. Generationary \cite{DBLP:conf/emnlp/BevilacquaMN20} leveraged BART to generate corresponding gloss for the span in contexts where the target word appears. 
\citet{DBLP:journals/corr/abs-2010-03124} proposed VCDM and introduced a continuous latent variable to explicitly model the underlying relationship between context and definition. 
Some works also studied other aspects of definition generation. \cite{DBLP:conf/emnlp/HuangKA21} utilized a re-ranking mechanism to model specificity in definitions. 
\citet{DBLP:conf/acl/KongCZYY22} emphasized simplification by proposing a SimpDefiner model which integrated three sub-tasks of definition generation with a weight sharing mechanism. Note that none of the aforementioned works attempted generating multiple definitions of a word at one time and exploiting coarse contexts possibly misaligned with the target definitions. 
Our work not only proposes a novel multiple definition modeling task but also constructs a new dataset for MDM with open-domain contexts.


\begin{table}[t]
\centering
\scalebox{0.85}{
\begin{tabular}{@{}l|l@{}}
\toprule[1.5pt]
\textbf{WORD}      & stable                                       \\ \midrule
\textbf{CONTEXT}   & a stable ladder                              \\ \midrule
\textbf{DEF}       & resistant to change of position or condition \\ \midrule
\textbf{SDM\_PRED} & capable of being shaped or bent or drawn out \\ \midrule
\textbf{MDM\_PRED} & free from collapse or unstable condition     \\ \midrule
\textbf{WORD}      & coddle                                       \\ \midrule[1.5pt]
\textbf{CONTEXT}   & coddle eggs                                  \\ \midrule
\textbf{DEF}       & to cook in nearly boiling water              \\ \midrule
\textbf{SDM\_PRED} & to wrap up in a tangle of eggs               \\ \midrule
\textbf{MDM\_PRED} & to cook on a hot surface using pressure      \\ \midrule
\textbf{WORD}      & competitive                                  \\ \midrule[1.5pt]
\textbf{CONTEXT}   & competitive games                            \\ \midrule
\textbf{DEF}       & involving competition or competitiveness     \\ \midrule
\textbf{SDM\_PRED} & involving much fun                           \\ \midrule
\textbf{MDM\_PRED} & involving or being in a competition          \\ \bottomrule[1.5pt]
\end{tabular}
}
\caption{An Example of predictions from T5-base trained with SDM and MDM task on WordNet dataset.}
\label{tab:case_study_app}
\end{table}

We select several cases in the predictions from both SDM-trained models and MDM-trained models, listed in \ref{tab:case_study_app}.

\end{document}